\documentclass[lettersize,journal]{IEEEtran}
\usepackage{amsmath,amsfonts,amssymb}
\usepackage{booktabs}
\usepackage{pifont}
\usepackage{graphicx}
\usepackage{caption}
\usepackage{subcaption}
\usepackage{url}
\usepackage{hyperref}
\usepackage{cite}

\begin{document}

\title{ELA: Exact Linear Attention with Qualitative Memory and Hyper-Link}

\author{Weinuo Ou,~\IEEEmembership{Student,~Wuyi University}
\thanks{Github: https://github.com/yauntyour/Exact-Linear-Attention}}

\markboth{Journal of \LaTeX\ Class Files,~Vol.~XX, No.~XX, XXXX~2026}%
{Ou: ELA---Exact Linear Attention with Qualitative Memory and Hyper-Link}
\maketitle

\begin{abstract}
The quadratic computational complexity of standard self-attention has driven a broad research effort into efficient alternatives spanning sparse patterns, locality-sensitive hashing, low-rank projections, kernelized approximations, and structured state-space recurrence. This paper presents \textbf{Exact Linear Attention (ELA)}, a mechanism that achieves $O(L)$ computational complexity while preserving mathematical exactness---departing fundamentally from prior approximate kernel methods (e.g., Performer, CosFormer, Linformer) that trade precision for efficiency, and from recurrent state-space models (e.g., Mamba, RWKV) that forgo explicit pairwise attention weights. By exploiting the exact decomposition property of kernel functions, ELA eliminates approximation error entirely while retaining the associative property that enables linear-complexity computation via the right-product trick. Crucially, ELA replaces the per-token KV cache with a fixed-size $O(1)$ accumulated state matrix---independent of sequence length---conferring decisive memory advantages at ultra-long contexts ($L > 100k$) where even optimized exact attention (FlashAttention) remains bottlenecked by $O(L)$ memory scaling.

We introduce a principled kernel design framework grounded in four criteria: exact decomposability, sufficient discriminability, non-negativity, and geometric interpretability. Three kernel families are derived under these criteria---the Hadamard Exp Kernel (feature co-activation patterns), the Summation Squared Euclidean Distance Kernel (directional alignment), and the Subtraction Squared Euclidean Distance Kernel (antagonistic relationships)---along with a blueprint for constructing task-specific custom kernels.

Beyond the core formulation, this paper advances three engineering contributions: (1) a \textbf{Hyper-Link} structure that rearchitects residual pathways by establishing cross-depth connections and eliminating the attention residual branch, mitigating gradient degradation more effectively than conventional residual connections; (2) a \textbf{Memory Lobe} module grounded in bidirectional linear attention over the layer-wise \textit{transformation flow} $\Delta X_{k|k-1}$, which realizes a mathematical formulation of qualitative memory and an implicit reinforcement learning paradigm via a natural Action--Reward mechanism---a novel paradigm for LLM training complementary to LoRA and Engram; and (3) a \textbf{routing-score-based bias mechanism} for Mixture-of-Experts (MoE) that couples expert label vectors to network outputs, enhancing semantic interpretability.

Benchmarks across language and vision domains demonstrate consistent gains. On language modeling, Hyper-Link delivers accelerated convergence and improved anti-overfitting; the Memory Lobe reduces convergence epochs from 30 to approximately 10. On vision tasks, YOLO-LAT achieves 2.2$\times$ faster CPU inference and 4.3$\times$ faster GPU inference with 7.9$\times$ fewer parameters, attaining competitive mAP@0.5 (0.962) while exposing a localization gap (mAP@0.5:0.95 of 0.515 vs.\ 0.951) attributable to the absence of explicit depth cues---a limitation that points toward depth-aware architectures as a natural extension. A simplified FCOS-based detector further confirms ELA's suitability for Vision Transformer workloads. All ELA operators are drop-in compatible with existing GPT-style models, establishing exact decomposability as a foundation for portable, scalable attention.
\end{abstract}

\begin{IEEEkeywords}
Exact Linear Attention, Kernel decomposition, Hadamard Exp Kernel, Summation Squared Euclidean Distance Kernel, Subtraction Squared Euclidean Distance Kernel, Linear complexity, Hyper-Link, Memory Lobe, Transformation Flow, Qualitative memory, KV cache efficiency, Inference throughput, Mixture-of-Experts (MoE), Routing-score bias, Long-sequence processing, Gradient explosion, Attention dilution, Object detection, Vision transformer, Reinforcement learning, Scaling laws, Model interpretability.
\end{IEEEkeywords}

\section{Introduction}
\IEEEPARstart{L}{inear} attention has long been regarded as a promising direction for scaling Transformers\cite{Vaswani2017} to long sequences. While standard self-attention delivers unparalleled representational power, its $O(L^2)$ time and memory complexity has spurred a broad research program into efficient alternatives. These span multiple algorithmic paradigms: \textbf{sparse attention} patterns that restrict each query to a subset of keys, as in Longformer’s sliding windows\cite{Beltagy2020longformer} and BigBird’s random+local+global composition\cite{Zaheer2020bigbird}; \textbf{locality-sensitive hashing (LSH)} that clusters similar keys, as in the Reformer\cite{Kitaev2020reformer}; \textbf{low-rank projections} that compress the key-value dimension, as in Linformer\cite{Wang2020linformer}; and \textbf{kernelized approximations} of the softmax, as in Performers\cite{Choromanski2021} and CosFormer\cite{Qin2022cos}. The Long Range Arena (LRA) benchmark\cite{Tay2021lra} established a standardized evaluation for these efficient attention models, revealing that while many perform competitively at moderate lengths, extreme long-range dependency modeling remains an open frontier.

Building upon the foundational \textbf{linear attention} framework—in which the kernel feature map is applied before the summation to achieve $O(L)$ complexity via the associative property—pioneered by Katharopoulos et al.\cite{Katharopoulos2020} and extended by Schlag, Irie, and Schmidhuber\cite{Schlag2021}, this paper formally presents an exact linear attention mechanism. Unlike prior kernel-based approximations such as Performers\cite{Choromanski2021} (which rely on random feature estimators of the softmax kernel), CosFormer\cite{Qin2022cos} (which substitutes ReLU-activated feature maps), or Linformer\cite{Wang2020linformer} (which projects the key-value sequence into a fixed low-rank subspace), the proposed approach exploits the exact decomposition property of kernel functions. This eliminates approximation error entirely while preserving the associative property that enables $O(L)$ computation.

Parallel to kernel-based linear attention, a distinct line of work has pursued linear-time sequence modeling through structured state-space recurrence. The Mamba architecture\cite{Gu2023mamba} introduced input-dependent selective state spaces that rival Transformers on language modeling while scaling linearly. RWKV\cite{Peng2023rwkv} reformulated the linear attention mechanism within a recurrent framework, achieving Transformer-level parallel training and RNN-level $O(1)$ inference. RetNet\cite{Sun2023retnet} similarly unifies parallel and recurrent representations through its multi-scale retention mechanism. These developments underscore the growing consensus that linear-complexity sequence modeling is both viable and necessary.

Subsequent work by the MiniMax Team systematically analyzed the limitations of linear attention and identified two critical issues: gradient explosion induced by denominator-driven unbounded gradients, and token attention dilution\cite{Qin2022}. In response to these deficiencies, we propose a linearized full attention computation framework that bounds gradients through carefully designed kernel constraints while leveraging the intrinsic favorable properties of kernels. This work offers a principled resolution to the long-standing limitations of linear attention.

By circumventing the $O(L^2)$ computational bottleneck of conventional attention—a bottleneck that even highly optimized exact attention implementations such as FlashAttention\cite{Dao2022flash} cannot eliminate at the asymptotic level—linear attention enables dramatic improvements in computational efficiency without recourse to sampling or sparsification. As the context length scales by an order of magnitude, the performance gap between linear and conventional attention widens proportionally. Linearized attention thus remains a vital research direction for Transformer-style architectures in the long term.

Beyond raw throughput, linear attention confers a decisive advantage in \textbf{KV cache efficiency}. Whereas standard multi-head attention (MHA) requires caching $2 \times L \times D \times n_{\text{heads}}$ key-value pairs, and even optimized variants such as Multi-Query Attention (MQA)\cite{Shazeer2019mqa} and Grouped-Query Attention (GQA)\cite{Ainslie2023gqa} only reduce this by a constant factor, linear attention replaces the per-token KV cache with a fixed-size accumulated state matrix. This state is independent of sequence length, fundamentally altering the memory scaling behavior.

Empirical evidence from leading industrial deployments confirms that this theoretical advantage translates into substantial performance gains in practice. Kimi Linear (The Dark Side of the Moon)\cite{kimilinear2025} and MiniMax\cite{minimax2025lightning} independently demonstrate:
\begin{itemize}
    \item \textbf{Inference Speed (Throughput).} For short texts (e.g., 4k context length), the speed difference between the two paradigms is marginal, and linear attention may even be marginally slower due to operator optimization immaturity. In ultra-long contexts (e.g., 1M tokens), however, the decoding speed of linear attention surpasses that of full global attention by more than sixfold. In a 1M-token benchmark, Kimi Linear achieves a Time Per Output Token (TPOT) of merely 1.84 ms, while traditional architectures such as MLA reach 11.48 ms.
    \item \textbf{Memory Usage (KV Cache).} Linear attention reduces the KV Cache memory footprint by 75\%–90\%.
\end{itemize}

These results imply that with identical GPU resources, linear attention models can process substantially longer documents or serve a greater number of concurrent users with larger batch sizes. As a case in point, MiniMax’s model leverages linear attention to support a context window of up to 4 million tokens---a scale that remains prohibitively expensive for conventional full-attention architectures.

Beyond raw efficiency, ELA occupies a distinct position in the design space of linear-complexity sequence models. State-space architectures such as Mamba~\cite{Gu2023mamba} and RWKV~\cite{Peng2023rwkv} achieve linear scaling by reformulating sequence mixing as structured recurrence; while effective, their latent state transitions are inherently implicit, making it difficult to inspect or interpret which tokens drive a given prediction. ELA, by contrast, preserves \textbf{explicit, interpretable pairwise attention weights} through exact kernel decomposition---retaining the transparency of the original attention mechanism while matching the asymptotic efficiency of SSMs. This property is critical not merely for interpretability, but for \textbf{architectural composability}: because ELA produces standard attention weight matrices (albeit computed in $O(L)$ time), it can serve as a drop-in module within any Transformer-based pipeline, including multi-modal fusion layers and world-model architectures, without requiring the bespoke training recipes that recurrent alternatives often demand.

Looking beyond the immediate efficiency gains, we regard ELA as a mathematical substrate for \textbf{general-purpose world models}. The exact decomposability of its kernel formulation, combined with the qualitative memory afforded by the Memory Lobe and the interpretable routing of MoE experts, provides a unified framework in which perception, memory, and reasoning coexist within a single attention-driven architecture. The remainder of this paper develops this framework in detail: \autoref{sec:kernel_criteria} establishes the kernel design principles, \autoref{sec:formulation} derives the exact linear attention formulation, \autoref{sec:engineering} presents the three engineering contributions, and \autoref{sec:experiments}--\ref{sec:vision} validate the approach on language and vision benchmarks.

\includegraphics[width=\linewidth]{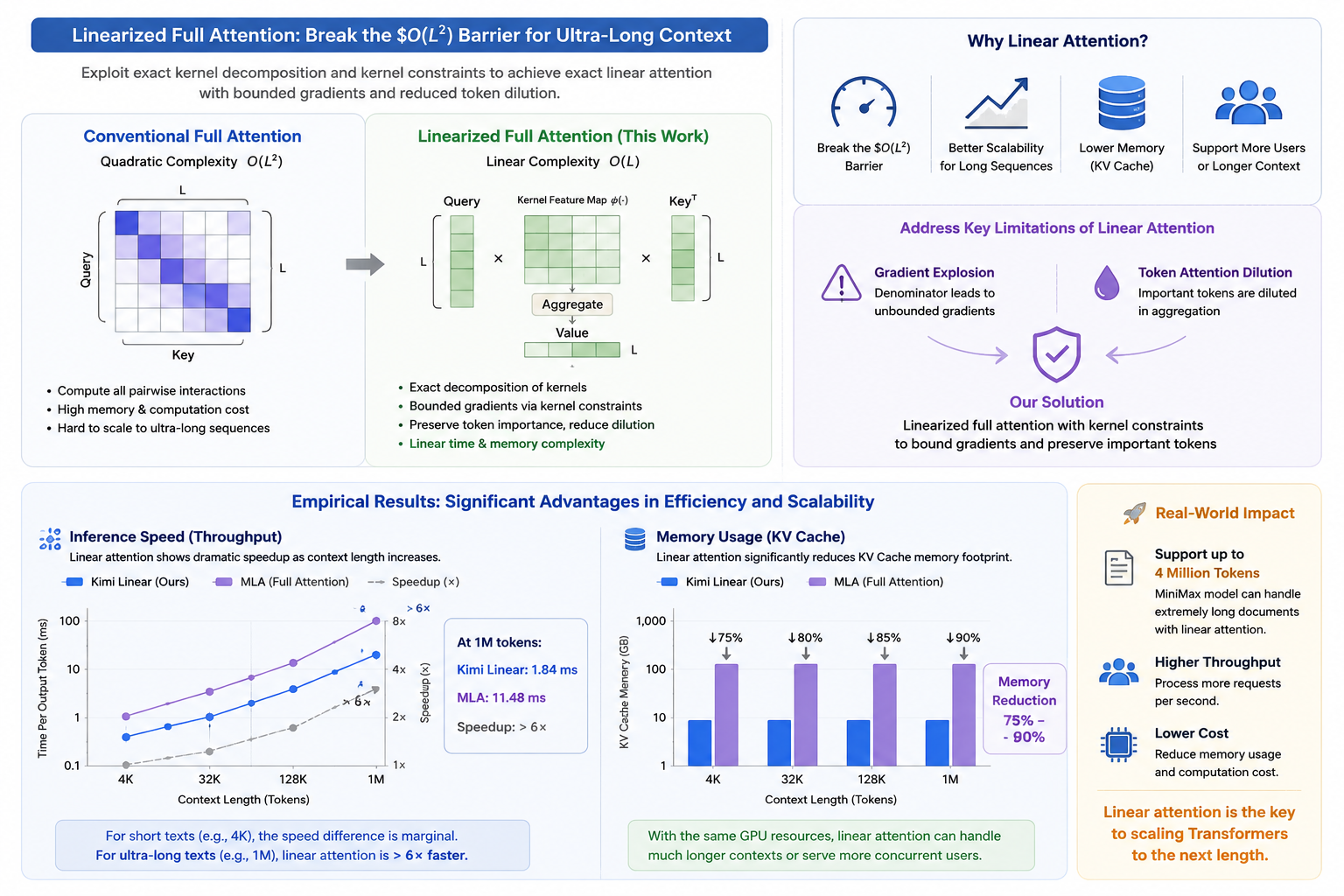}

\section{Kernel Selection Criteria for Exact Linear Attention}
\label{sec:kernel_criteria}
To systematically address the inherent limitations of linear attention, we first enumerate the desirable properties that an ideal kernel function ought to satisfy. \autoref{tab:kernel_criteria} presents the four design criteria that jointly govern kernel selection throughout this work.
\begin{table}[htbp]
\centering
\caption{Kernel design criteria for Exact Linear Attention.}
\label{tab:kernel_criteria}
\begin{tabular}{@{}p{0.28\columnwidth}p{0.62\columnwidth}@{}}
\toprule
\textbf{Criterion} & \textbf{Requirement} \\
\midrule
Exactly decomposable & The kernel must admit a closed-form factorization $k(A_i,B_j)=\phi(A_i)\psi(B_j)^\top$ in a finite-dimensional feature space, enabling exact $O(L)$ computation via the associative property. \\
\addlinespace
Sufficiently discriminative & The output curve must exhibit sharp discriminability with a smooth and bounded value range, thereby mitigating gradient vanishing and explosion during training. \\
\addlinespace
Non‑negative & $k(A_i,B_j) \ge 0$ for all input pairs, ensuring that all attention weights form a valid probability distribution after row-wise normalization. \\
\addlinespace
Geometrically interpretable & The kernel should admit a clear geometric interpretation in the embedding space, enhancing the internal transparency and debuggability of the attention mechanism. \\
\bottomrule
\end{tabular}
\end{table}
These four criteria encapsulate the essential characteristics required for principled attention computation. Exact decomposability guarantees numerical precision; sufficient discriminability prevents the attention matrix from being diluted after row-wise normalization; non-negativity and geometric interpretability together establish the foundational prerequisites for transparent, interpretable attention modeling.

Through systematic categorization and analysis, we find that kernel functions satisfying these requirements broadly fall into the following families: polynomial-type, exponential-type, non-negative periodic function-type, and absolute value function-type.

Among these candidates, the Hadamard Exp Kernel emerges as a particularly compelling choice. \autoref{tab:hadamard_advantages} summarizes its key advantages relative to alternative kernel families.
\begin{table}[htbp]
\centering
\caption{Advantages of the Hadamard Exp Kernel over alternative kernel candidates.}
\label{tab:hadamard_advantages}
\begin{tabular}{@{}p{0.32\columnwidth}p{0.58\columnwidth}@{}}
\toprule
\textbf{Property} & \textbf{Description} \\
\midrule
Exact decomposability & Admits a closed-form factorization $\phi(A_i)^\top\psi(B_j)$ in $\mathbb{R}^D$ without requiring infinite series expansion or stochastic approximation. \\
\addlinespace
Strong nonlinearity & The element-wise exponential transformation $\exp(\cdot)$ amplifies co-activated feature pairs while suppressing low-intensity noise, providing sharper feature selectivity than linear or polynomial kernels. \\
\addlinespace
Smoothness & Infinitely differentiable everywhere ($C^\infty$), ensuring well-behaved gradients throughout the entire input domain and stable optimization dynamics. \\
\addlinespace
Non-negativity & $\exp(x) > 0$ for all real $x$, guaranteeing strictly positive attention weights without additional activation constraints. \\
\bottomrule
\end{tabular}
\end{table}
These properties naturally connect to the essence of attention—query-based cosine-like similarity—and illuminate the distinct geometric interpretability of the three kernel families proposed in this work. \autoref{tab:kernel_comparison} provides a comparative overview.
\begin{table}[htbp]
\centering
\caption{Comparison of the three proposed kernel families.}
\label{tab:kernel_comparison}
\begin{tabular}{@{}p{0.26\columnwidth}p{0.28\columnwidth}p{0.36\columnwidth}@{}}
\toprule
\textbf{Kernel} & \textbf{Attention Mechanism} & \textbf{Typical Application} \\
\midrule
Summation Squared Euclidean Distance
$\|A_i+B_j\|^2$
& Emphasizes keys aligned in the \textbf{same} direction as the query; retrieves supporting evidence consistent with query semantics.
& Question answering, document retrieval, paraphrase identification. \\
\addlinespace
Subtraction Squared Euclidean Distance
$\|A_i-B_j\|^2$
& Emphasizes keys \textbf{opposite} or antagonistic to the query; captures contrastive relationships.
& Contrastive learning, contradictory paragraph detection, anomaly detection. \\
\addlinespace
Hadamard Exp
$\exp(A_i)\!\ast\!\exp(B_j)$
& Emphasizes \textbf{feature co-activation} patterns via exponential amplification; suppresses low-intensity noise while enhancing salient signals.
& Multimodal fusion, feature gating, noise-robust representation learning. \\
\bottomrule
\end{tabular}
\end{table}
While most other kernel variants are derivatives of these three archetypes, the \textbf{Hadamard Exp Kernel} merits particular attention relative to canonical attention. Its element-wise exponential product characterizes the co-occurrence intensity across feature dimensions: the exponential transformation naturally amplifies strongly activated feature pairs while suppressing low-intensity noise. Compared with the cosine-similarity-based paradigm of conventional attention\cite{Vaswani2017}, the Hadamard Exp Kernel captures feature co-activation patterns with greater discriminability, offering clear advantages in scenarios demanding fine-grained feature interaction. Analogous benefits hold for the summation and subtraction Euclidean distance kernels, which encode magnitude–directional relationships. The Hadamard Exp Kernel, however, is especially well-suited to multimodal settings where cross-modal semantic feature co-activation is critical.

It is worth noting that prior linear attention approaches with kernel-based feature maps—such as Performer's positive random features\cite{Choromanski2021} and CosFormer's ReLU-based decomposition\cite{Qin2022cos}—either rely on stochastic approximations or sacrifice exactness for efficiency. In contrast, our kernel formulations preserve mathematical exactness while retaining full compatibility with the associative property that enables linear-complexity computation.

\section{Exact Linear Attention Formulation}
\label{sec:formulation}
We now derive the attention formula following the standard attention paradigm.
Let $A \in \mathbb{R}^{B \times L \times D}$ and $B \in \mathbb{R}^{B \times L \times D}$ be the query and key representations, and $V \in \mathbb{R}^{B \times L \times d_v}$ the value matrix. We index individual token vectors as $A_i, B_j \in \mathbb{R}^{D}$ and $V_j \in \mathbb{R}^{d_v}$, where $i, j \in \{1,\ldots,L\}$ and $B$ denotes the batch dimension.

By virtue of Mercer's theorem\cite{mercer1909}, any positive definite kernel admits a decomposition as an inner product within a feature space.
\begin{align}
& k(A_i, B_j) = \sum_{m=1}^{\infty}\lambda_m \phi_m(A_i)\psi_m(B_j)^\top \label{eq:decomp} 
\end{align}

However, we do not require such a fully positive definite decomposition property here. It is sufficient for the kernel function operation on $A_i$ and $B_j$ to be decomposed into the product of two sub-kernels.
\begin{align}
& k(A_i, B_j) = \phi(A_i)\psi(B_j)^\top
\end{align}
The decomposition of the aforementioned kernel functions can be illustrated as follows:
\begin{itemize}
    \item \textbf{Summation Squared Euclidean Distance Kernel}
    \begin{align}
        & k(A_i,B_j) = \|A_i + B_j\|^2 = \|A_i\|^2 + \|B_j\|^2 + 2A_i\cdot B_j \nonumber\\
        & \phi(A_i) = \begin{pmatrix} A_i \\ \|A_i\|^2 \\ 1 \end{pmatrix} \in \mathbb{R}^{D+2},\nonumber\\
        & \psi(B_j) = \begin{pmatrix} 2B_j \\ 1 \\ \|B_j\|^2 \end{pmatrix} \in \mathbb{R}^{D+2}.
    \end{align}
    \item \textbf{Subtraction Squared Euclidean Distance Kernel}
    \begin{align}
        & k(A_i,B_j) = \|A_i - B_j\|^2 = \|A_i\|^2 + \|B_j\|^2 - 2A_i\cdot B_j \nonumber\\
        & \phi(A_i) = \begin{pmatrix} A_i \\ \|A_i\|^2 \\ 1 \end{pmatrix} \in \mathbb{R}^{D+2},\nonumber\\
        & \psi(B_j) = \begin{pmatrix} -2B_j \\ 1 \\ \|B_j\|^2 \end{pmatrix} \in \mathbb{R}^{D+2}.
    \end{align}
    \item \textbf{Hadamard Exp Kernel}
    \begin{align}
        & k(A_i,B_j) = \exp(A_i) \ast \exp(B_j) = \sum_{d=1}^{D} \exp(A_{id})\exp(B_{jd}) \nonumber\\
        & \phi(A_i) = \begin{pmatrix} \exp(A_{i1}) \\ \vdots \\ \exp(A_{iD}) \end{pmatrix} \in \mathbb{R}^{D},\quad \nonumber\\
        & \psi(B_j) = \begin{pmatrix} \exp(B_{j1}) \\ \vdots \\ \exp(B_{jD}) \end{pmatrix} \in \mathbb{R}^{D}.
    \end{align}
\end{itemize}

For further illustration, note that exact decomposition imposes only weak symmetry requirements on the kernel.
\begin{align}
    &k(A_i, B_j) = A_{i1} B_{j2} + 2 A_{i2} B_{j1} \nonumber\\
    &\phi(A_i) = \begin{pmatrix} A_{i1} \\ 2A_{i2} \end{pmatrix} \in \mathbb{R}^{2},\quad
    \psi(B_j) = \begin{pmatrix} B_{j2} \\ B_{j1} \end{pmatrix} \in \mathbb{R}^{2}.
\end{align}
Clearly, $k(A_i, B_j) \neq k(B_j, A_i)$ in general, showing that the decomposition $k(A,B)=\langle\phi(A),\psi(B)\rangle$ does not require the kernel to be symmetric.

Crucially, the factorization $k(A_i, B_j) = \phi(A_i)\,\psi(B_j)^\top$ cleanly separates the query-dependent and key-dependent computations. This separation enables the associative reordering that yields linear-complexity attention \textbf{without any loss of precision}.
\begin{align}
k(A_i, B_j)V_j = \phi(A_i) \psi(B_j)^\top V_j = \phi(A_i)\,[\psi(B_j)^\top V_j]
\end{align}
Building on this factorization, we apply row-wise normalization to $k(A_i, B_j)$ so that each row forms a valid probability distribution. This normalizes the attention weights while eliminating the softmax operation entirely. When combined with the summation reordering described below, the entire pipeline retains $O(L)$ computational complexity.

\begin{align}
\frac{\sum_{j=1}^{L} k(A_i, B_j)V_j}{\sum_{j=1}^{L} k(A_i, B_j)} &= \frac{\phi(A_i)\sum_{j=1}^{L}\psi(B_j)^\top V_j}{\sum_{j=1}^{L} \phi(A_i)\psi(B_j)^\top} \nonumber\\
&= \frac{\phi(A_i)\bigl[\sum_{j=1}^{L}\psi(B_j)^\top V_j\bigr]}{\phi(A_i)\sum_{j=1}^{L}\psi(B_j)^\top}
\end{align}

In practice, we adopt the following accumulation strategies to implement bidirectional and causal linear attention efficiently:
\begin{itemize}
    \item \textbf{Bidirectional Attention}
        \begin{align}
        C &= \sum_{j=1}^{L} \psi(B_j), \qquad
        S = \sum_{j=1}^{L} \psi(B_j) V_j^\top, \label{eq:bidir_context} \\
        Y_i &= \frac{\phi(A_i)^\top S}{\phi(A_i)^\top C}. \label{eq:bidir_out}
        \end{align}
    \item \textbf{Causal (Auto-Regressive) Attention}
        \begin{align}
        C_i &= \sum_{j=1}^{i} \psi(B_j), \qquad
        S_i = \sum_{j=1}^{i} \psi(B_j) V_j^\top, \label{eq:causal_context} \\
        Y_i &= \frac{\phi(A_i)^\top S_i}{\phi(A_i)^\top C_i}. \label{eq:causal_out}
        \end{align}
\end{itemize}
By swapping the order of summation, the bidirectional version requires only a single accumulation over the sequence, and the causal version uses a prefix sum (cumulative sum). In both cases the entire attention output is computed in $O(L)$ time without ever materializing the $L \times L$ attention matrix.  
Because the kernel is exactly decomposable into finite‑dimensional feature maps, the result is mathematically identical to the full quadratic form—this is an \textbf{exact}, rather than approximate, linear attention mechanism.

\section{Designing Custom Attention Kernels}
The four criteria established in the preceding section provide a principled blueprint for constructing task-specific attention kernels. Any kernel function satisfying these four requirements can be linearized to achieve $O(L)$ time complexity without approximation—a degree of flexibility that invites readers to design custom kernels tailored to their own applications.

As a concrete illustration, consider the goal of approximating standard softmax attention with maximal fidelity. Standard attention can be interpreted as computing a scaled dot product, applying an exponential transformation, and then normalizing—a pipeline equivalent to the exponential dot-product kernel. Evaluating this kernel in exact decomposed form, however, would necessitate a truncated Taylor expansion. We therefore pursue an alternative: constructing a specialized kernel that satisfies the exact decomposition condition while preserving the characteristic ability of standard attention to capture both vector magnitude and directional information simultaneously.

The resulting construction takes the following form:
\begin{align}
& k(A_i, B_j) = (\vec{A}_i \cdot \vec{B}_j + 1) \cdot (\|A_i\|^2 + 1) \cdot (\|B_j\|^2 + 1) \nonumber\\
& \phi(A_i) = (\|A_i\|^2 + 1)\begin{pmatrix} \vec{A}_i \\ 1 \end{pmatrix}, \quad \psi(B_j) = (\|B_j\|^2 + 1)\begin{pmatrix} \vec{B}_j \\ 1 \end{pmatrix} \nonumber\\
& \phi(A_i)^\top \psi(B_j) = (\|A_i\|^2 + 1)(\|B_j\|^2 + 1)(\vec{A}_i \cdot \vec{B}_j + 1)
\end{align}
Although its algebraic form may appear elaborate, this kernel admits a clean decomposition into two semantically meaningful components. The term $\vec{A}_i \cdot \vec{B}_j + 1$ captures directional attention, while $(\|A_i\|^2 + 1) \cdot (\|B_j\|^2 + 1)$ encodes magnitude-based attention. Following this compositional paradigm, one can, in principle, construct an arbitrarily rich family of attention kernel functions.

\section{Engineering Challenges and Solutions}
\label{sec:engineering}
Modern machine learning frameworks such as PyTorch have made rapid prototyping of model architectures straightforward. Nevertheless, the path from prototype to AGI-capable systems remains obstructed by communication overhead, memory consumption, energy cost, and even organizational factors. We contend that these challenges are surmountable through technological advances that liberate engineering productivity. In the following, we examine several key engineering bottlenecks and propose corresponding solutions.

\subsection{The FFN Interpretability Dilemma}
A central limitation of Feed-Forward Networks (FFNs) is their poor interpretability—the well-known "\textbf{black-box}" problem\cite{jain2019attention}. Existing approaches to tracing FFN computations predominantly rely on post-hoc statistical characterization of pre-trained models. Such methods, however, are largely ineffectual for Mixture-of-Experts (MoE) architectures. The pioneering work on sparsely-gated MoE\cite{Shazeer2017} and its integration into Transformers via GShard\cite{Lepikhin2020} and Switch Transformers\cite{Fedus2021} have demonstrated the scalability of conditional computation. However, the sparse activation inherent to MoE introduces a gradient inconsistency gap between the router and the expert groups. Under hard-constrained load balancing, each expert is trained in near isolation, rendering it intractable to explain why a given set of tokens activates a particular expert. Attributing expert selection merely to superior processing capability for a specific token type is, at best, tenuous and lacks broad acceptance. Furthermore, token-level routing dispatch imposes substantial communication overhead\cite{Fedus2021}. While we fully acknowledge the significance of MoE for expanding the knowledge capacity of neural networks, we seek an alternative: a method that performs nonlinear transformations on post-attention semantic embeddings without relying on explicit per-token routing.

An attention-based query mechanism is a natural candidate. Although traditional full attention has long been avoided due to its prohibitive computational cost, the advent of linear-complexity attention fundamentally changes this trade-off. We assign each expert network a fixed, learnable "\textbf{label vector}"; these vectors are aggregated into a unified key representation during computation, analogous to multi-head attention. The resulting workflow is conceptually clean: we query the expert label vectors within the semantic space, and experts exhibiting high semantic co-occurrence are identified as possessing knowledge most relevant to the query.

A critical question then arises: how can we ensure that these label vectors faithfully represent the capabilities of their corresponding experts? In other words, the model requires an intrinsic mechanism that couples label vectors to the output behavior of each expert. \autoref{tab:label_vector_methods} presents two simple yet effective coupling strategies, neither requiring complex auxiliary mappings.
\begin{table}[htbp]
\centering
\caption{Strategies for coupling expert label vectors to network outputs.}
\label{tab:label_vector_methods}
\begin{tabular}{@{}p{0.22\columnwidth}p{0.68\columnwidth}@{}}
\toprule
\textbf{Method} & \textbf{Mechanism} \\
\midrule
Bias injection & Treat the label vector directly as the bias term of the expert network, establishing an additive coupling between the semantic label and the expert's output activation. \\
\addlinespace
Weight factorization & Map the label vector into a component of the expert's weight matrix via low-rank factorization $W = W_0 + u\,v^\top$, where $v$ is derived from the label embedding. \\
\bottomrule
\end{tabular}
\end{table}

In most MoE implementations, the routing score serves as the fusion weight across multiple experts. One may interpret the weighted summation of routing scores as a transformation operation over vectors in the embedding space—an implicit form of internal semantic transformation. This process bears a loose analogy to the cognitive act of "association" that humans routinely perform. Human association, however, is intrinsically attention-aware: attention pervades the entire cognitive process\cite{buschman2010goal}, a level of integration that current AI struggles to approach, whether due to theoretical limitations or hardware constraints. Achieving human-like associative reasoning may ultimately require computational substrates—such as quantum-state computing—capable of superposing attention distributions across multiple possibility branches simultaneously.

\begin{figure}[htbp]
    \centering

    \textbf{Exact Linear Attention GPT}\\
    \begin{subfigure}{0.32\columnwidth}
        \centering
        \caption{Inner bias}
        \includegraphics[width=\textwidth]{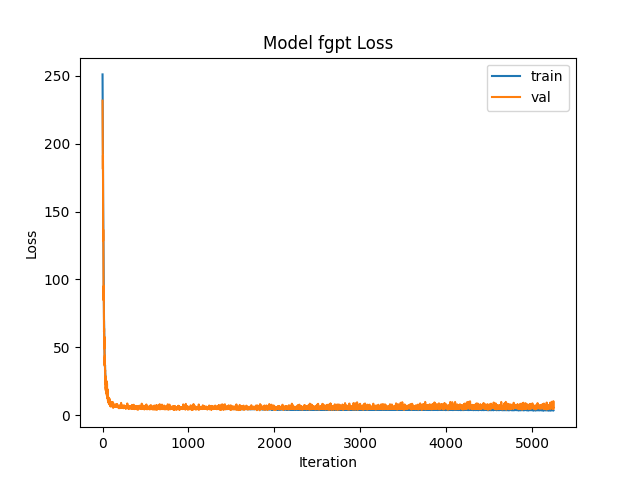}
    \end{subfigure}
    \hfill
    \begin{subfigure}{0.32\columnwidth}
        \centering
        \caption{Outer bias}
        \includegraphics[width=\textwidth]{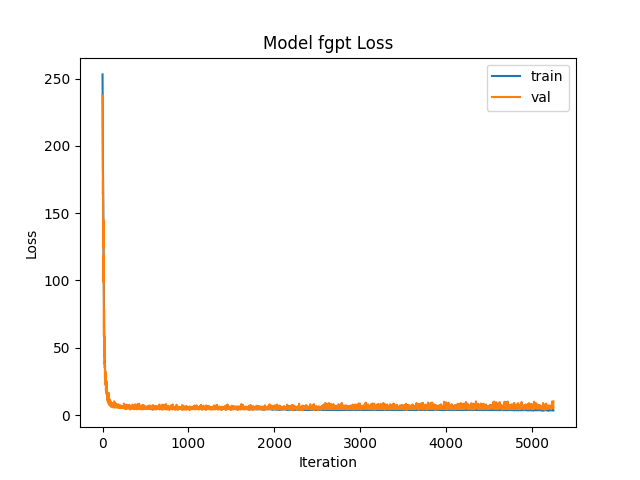}
    \end{subfigure}
    \hfill
    \begin{subfigure}{0.32\columnwidth}
        \centering
        \caption{Without bias}
        \includegraphics[width=\textwidth]{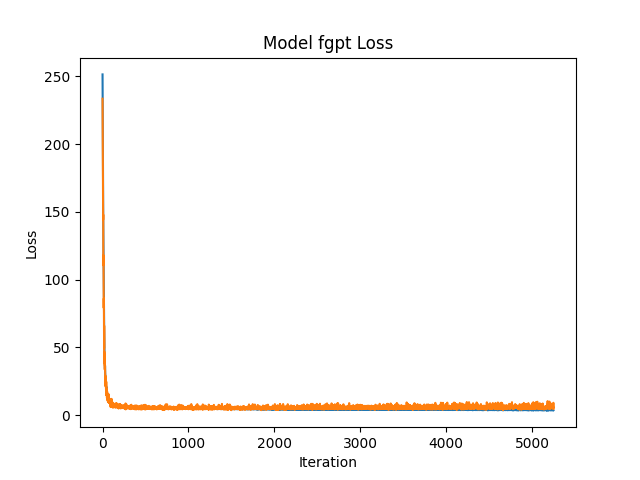}
    \end{subfigure}

    \textbf{Full Attention GPT}\\
    \begin{subfigure}{0.32\columnwidth}
        \centering
        \caption{Inner bias}
        \includegraphics[width=\textwidth]{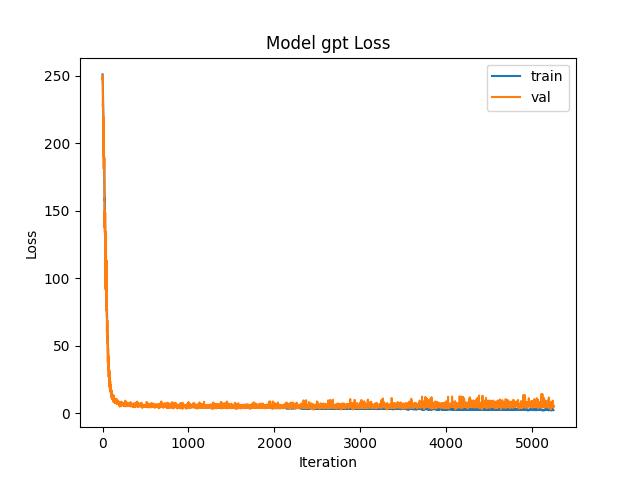}
    \end{subfigure}
    \hfill
    \begin{subfigure}{0.32\columnwidth}
        \centering
        \caption{Outer bias}
        \includegraphics[width=\textwidth]{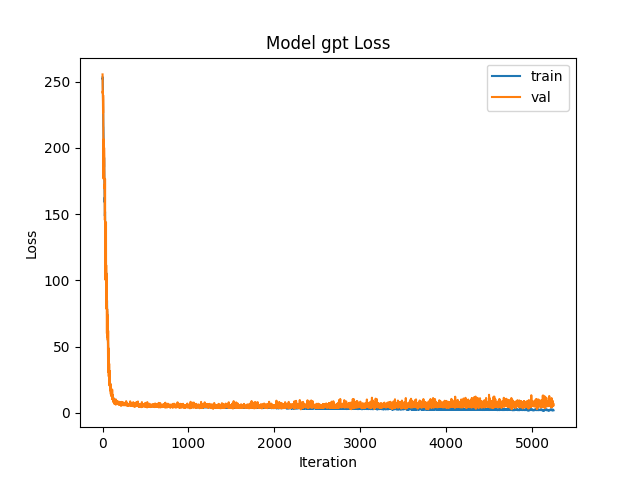}
    \end{subfigure}
    \hfill
    \begin{subfigure}{0.32\columnwidth}
        \centering
        \caption{Without bias}
        \includegraphics[width=\textwidth]{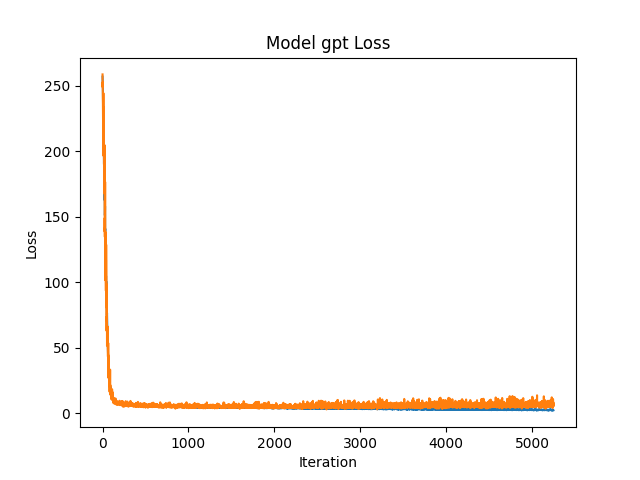}
    \end{subfigure}
    \caption{Comparison of Exact Linear Attention GPT (top row) and Full Attention GPT (bottom row).}
    \label{fig:ffns}
\end{figure}

Achieving interpretability for FFNs ultimately requires characterizing their transformation dynamics in terms of the representational manifold of the embedding space. Two concrete formulations for incorporating routing weights as bias terms are given below:
\begin{align}
    & X_t = S_e * ffn(X_{t-1})+B_e \\
    & X_t = S_e * (ffn(X_{t-1}) + B_e)
\end{align}
The distinction between these two formulations is whether the gradient propagates through the routing score: their differential matrices differ only by a multiplicative factor of the routing score. Comparative experiments (\autoref{fig:ffns}) demonstrate that this discrepancy is empirically negligible. Regardless of the bias variant adopted, however, both consistently outperform the bias-free baseline.

More elaborate mapping schemes warrant further investigation. For the present work, we focus on the current MoE architecture as a case study, demonstrating that sliced routing-score weights used as bias terms improve the semantic alignment between input and output transformations.
\includegraphics[width=\linewidth]{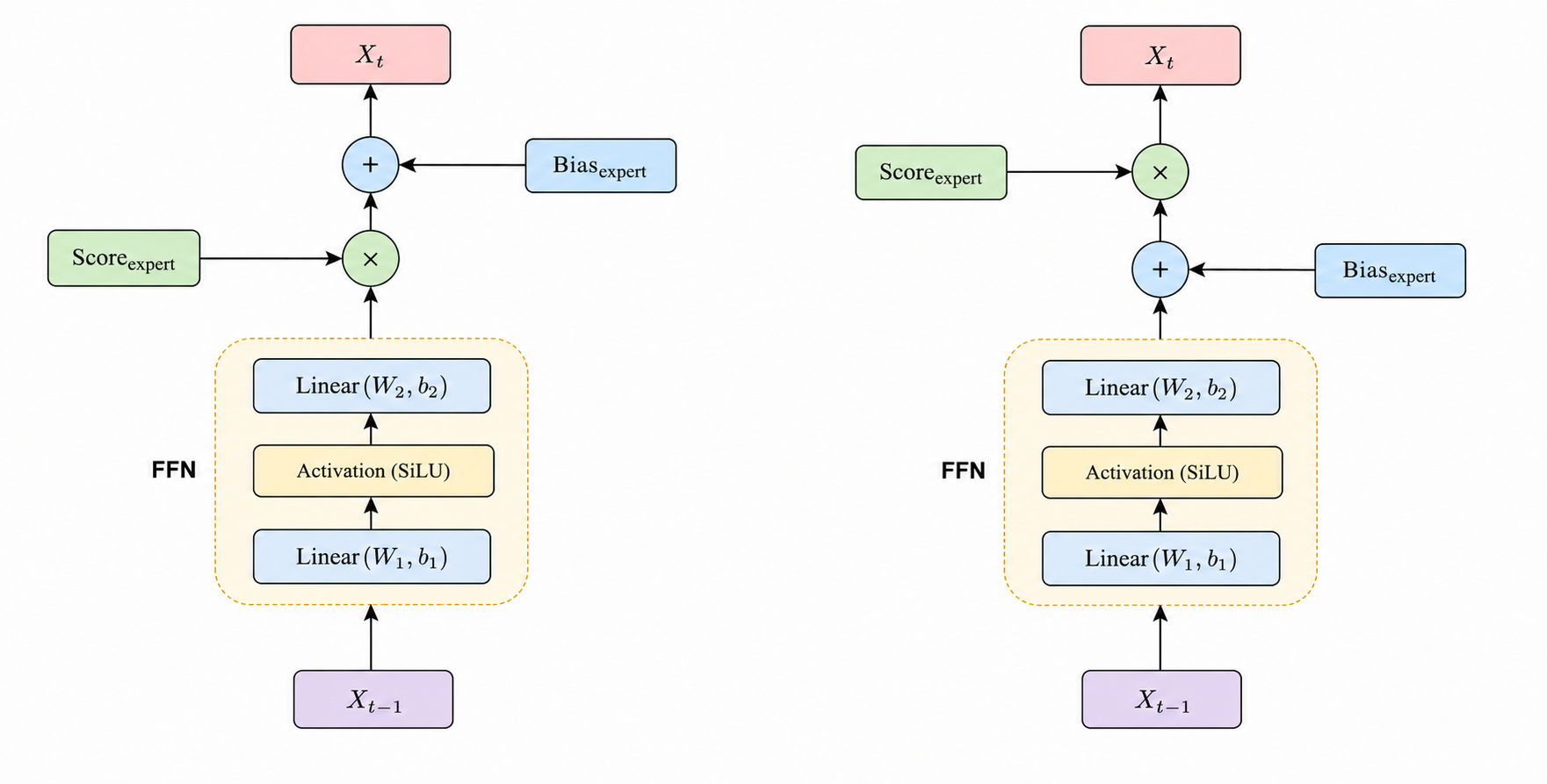}

Regarding communication overhead, token dispatching via routing scores remains unavoidable under sparse activation. Cross-device transmission can, however, be scheduled in a unified fashion analogous to the unified memory architecture design\cite{jia2020megatron}. A key structural observation is that consecutive token blocks naturally form semantic communities. \textbf{Partitioning tokens into semantically coherent blocks—rather than routing at the individual token level—can substantially reduce communication overhead.}

\subsection{Hyper-Link: Residual Replacement for Gradient Degradation Mitigation}
Traditional residual connections\cite{He2016resnet} across deep Decoder stacks suffer from gradient vanishing and impeded cross-layer information propagation. While Pre-LN architectures\cite{Xiong2020} mitigate training instability relative to the original Post-LN design, they do not fundamentally address degradation across extreme depths. The current state-of-the-art solutions to this problem are Hyper-Connections (HC)\cite{zhu2024hyper} and Manifold-Constrained Hyper-Connections (mHC)\cite{xie2025mhc}.

We propose a more fundamental intervention: reconstructing the residual pathway itself. Specifically, we establish direct residual connections between Decoder layers at different depths while removing the attention residual branch in the standard Pre-Norm architecture\cite{Xiong2020}, thereby treating the entire Transformer layer as an integrated computational unit. Furthermore, since modern FFNs are equipped with gated linear units, their gated outputs can be naturally harnessed to adaptively modulate each layer's signal contribution.

\includegraphics[width=\columnwidth]{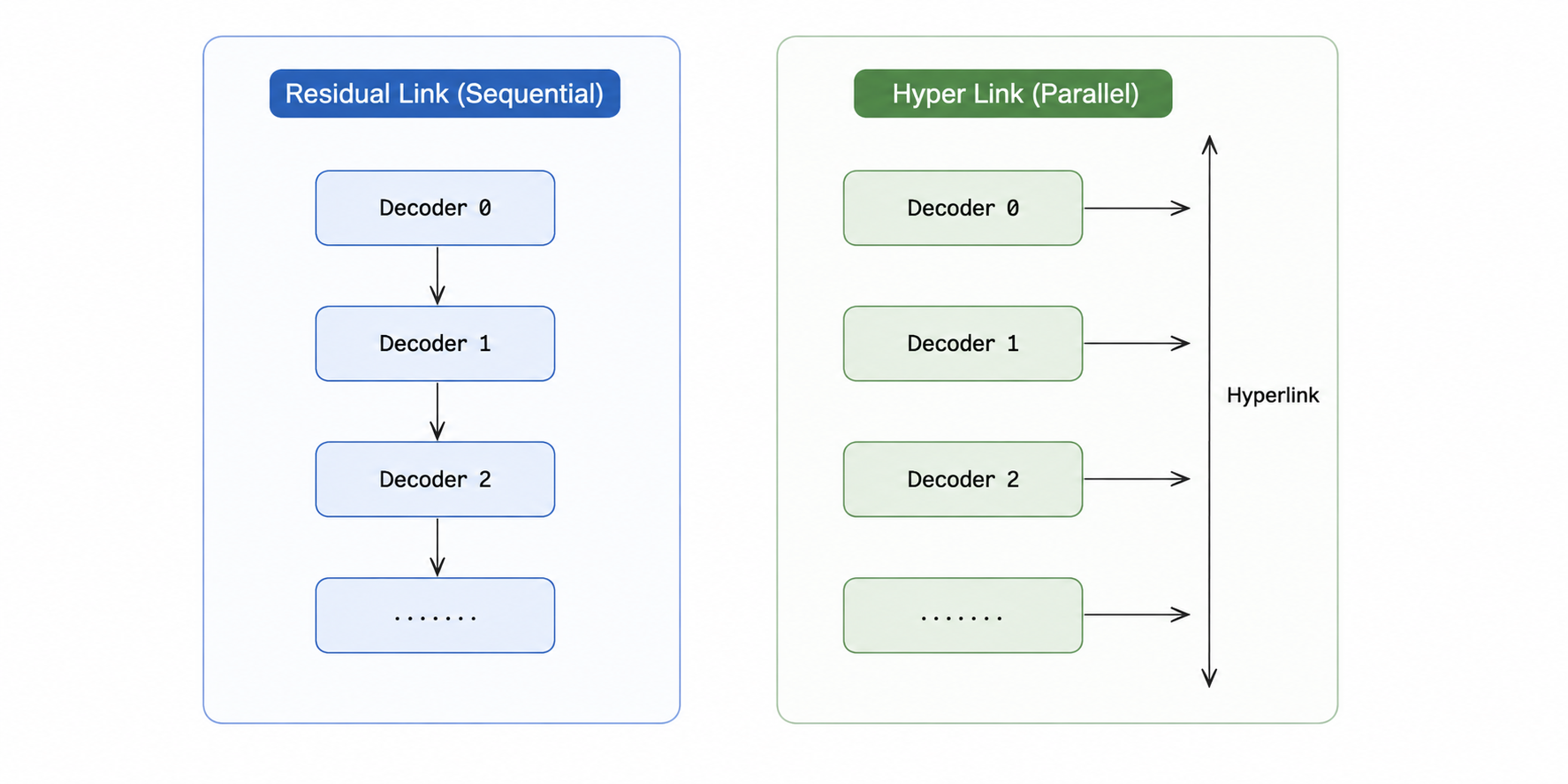}

Experimental results demonstrate that Hyper-Link effectively accelerates training and substantially mitigates gradient degradation. Under identical computational budgets, Hyper-Link achieves faster convergence and superior fitting performance relative to conventional residual connections. This mechanism also accounts for the characteristic convergence profile observed across all experimental plots—an extremely steep initial loss drop followed by a smooth, steady decline in later stages.

\textbf{For rapid prototyping, we removed GPT's final normalization layer to accelerate convergence.} In production-scale cluster training, however, the final normalization layer is necessary to preserve model stability. For gated hyper-link connections, output normalization is redundant since the gate inherently modulates the output signal. For extremely large and semantically diverse datasets—corpora spanning multiple domains with high token-level semantic entropy—additional segmented normalization may be required to sustain training stability.
\begin{figure}[htbp]
    \centering
    \begin{subfigure}{0.48\columnwidth}
        \centering
        \caption{Hyper-Link}
        \includegraphics[width=\textwidth]{./assets/gpt_loss.png}
    \end{subfigure}
    \hfill
    \begin{subfigure}{0.48\columnwidth}
        \centering
        \caption{Normal}
        \includegraphics[width=\textwidth]{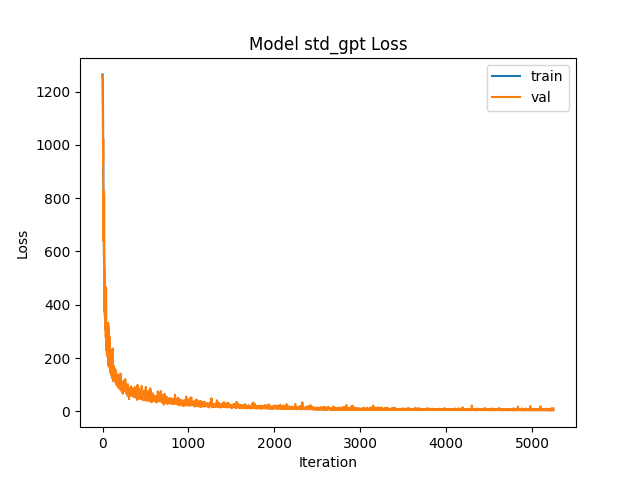}
    \end{subfigure}

    \caption{Training Comparison (GPT)}
    \label{fig:ffn_comp}
\end{figure}

\subsection{Qualitative Memory via Transformation Flow}
Human memory can be broadly dichotomized into two forms. The first, which we term \textbf{factual memory}, records the occurrence of events. The second, \textbf{qualitative memory}, encodes how events are perceived or evaluated. This fundamental distinction partitions all known information into two categories: objective existence and behavioral judgment.

Factual memory functions as background knowledge—a repository of what has transpired. Qualitative memory, in contrast, operates more like an intrinsic system of constraints and rules that shape future behavior. The distinction is readily illustrated: if one has a disappointing dining experience at a particular restaurant, the decision of whether to return is guided not by the factual record of the visit, but by the \textbf{evaluative content} embedded in qualitative memory.

In conventional model training, both cognitive capabilities are entirely subsumed within the Feed-Forward Network (FFN), producing an opaque black box in which multiple conditional constraints are tightly entangled. Disentangling factual from qualitative memory therefore demands a redesign of the computational process from first principles—specifically, from the perspective of semantic space transformation.

Recent work by the DeepSeek team demonstrates that the Engram module\cite{cheng2026conditional} performs effectively as an auxiliary knowledge storage component, corresponding to factual memory. This naturally raises the complementary question: how should we construct qualitative memory, the more critical substrate of behavioral guidance? Our answer echoes a familiar maxim: \textbf{Attention is all you need.}

Recall that in the Hyper-Link design, we deliberately \textbf{removed the attention residual branch}. This decision serves a dual purpose: beyond enabling the layer output to function as a cohesive whole, it unlocks an elegant application for memory modeling.
If we compute the discrete differential of the layer transformation $X_{k} \to X_{k-1}$, we obtain:
\begin{align}
X_{k} &= DecoderLayer(X_{k-1}) \nonumber\\
\Delta X_{k|k-1} &= X_{k} - X_{k-1} \nonumber\\
&= ffn(attn(RMSnorm(X_{k-1})\,)\,)\nonumber
\end{align}
We denote the differential $\Delta X_{k|k-1}$ as the \textbf{Transformation Flow} of the mapping $X_{k} \to X_{k-1}$. It captures the trajectory of semantic evolution through the current layer—an object that records how representations are transformed. We then design a bidirectional attention-based perception module that operates over this Transformation Flow, formulated through the learned projections summarized in \autoref{tab:flow_projections}.
\begin{table}[htbp]
\centering
\caption{Learned projections of the Transformation Flow perception module.}
\label{tab:flow_projections}
\begin{tabular}{@{}c p{0.22\columnwidth} p{0.45\columnwidth}@{}}
\toprule
\textbf{Matrix} & \textbf{Role} & \textbf{Semantic Interpretation} \\
\midrule
$Q \in \mathbb{R}^{D \times D}$ & Query & ``What to attend to in this transformation'' — the pattern the memory module seeks to recognize in the flow. \\
\addlinespace
$K \in \mathbb{R}^{D \times D}$ & Key & ``What this transformation provides'' — the indexable signature of the current layer's processing behavior. \\
\addlinespace
$V \in \mathbb{R}^{D \times D}$ & Value & ``What this transformation contributes'' — the information payload carried by the flow for downstream generation. \\
\bottomrule
\end{tabular}
\end{table}
Pseudocode:
\begin{verbatim}
def lob(dx):
    q = Q(dx)
    k = K(dx)
    v = V(dx)
    # Bidirectional Linear Attention
    return ELA(q, k, v)
...
def decoder(x):
    x_norm = norm(x)
    attn = ELA_causal(query=x_norm,
        key=x_norm,
        value=x_norm)
    ffn_out, aux_loss = MoE(attn)
    # get the flow query attention output
    lob_out = lob(ffn_out)
    # hyper-link
    return x + ffn_out + lob_out, aux_loss
\end{verbatim}
{\footnotesize  For detailed implementation, please refer to our GitHub repository.}

With this construction, the DecoderLayer augmented with Transformation Flow comprehensively outperforms the vanilla variant during training. Datasets that originally required 30 epochs for convergence reach comparable performance in approximately 10 epochs once the memory module is integrated, with substantially improved alignment between training and validation loss curves.

This architecture constitutes a mathematical realization of \textbf{qualitative memory}. The QKV weight matrices of the bidirectional attention module learn to encode which representational transformations lead to lower loss during training. Since the input to the module is the \textbf{layer-wise Transformation Flow} itself, the model implicitly accumulates layer-specific processing experience that informs subsequent generation.

Because the FFN output derives from causal attention, it inherently possesses forward causal properties. The memory query, however, requires access to the transformation history across all positions—hence the adoption of global bidirectional attention for this component.

Although the training procedure superficially resembles standard supervised learning, it effectively implements an implicit reinforcement learning paradigm. The entire pipeline depends on parameterized memory content—queried from layer Transformations—to support the final output, forming a natural "Action–Reward" mechanism: the current memory query constitutes the \textbf{Action}, and its direct contribution to the loss function serves as the corresponding \textbf{Reward}.

\begin{figure}[htbp]
    \centering
    \begin{subfigure}{0.48\columnwidth}
        \centering
        \caption{Memory lobe}
        \includegraphics[width=\textwidth]{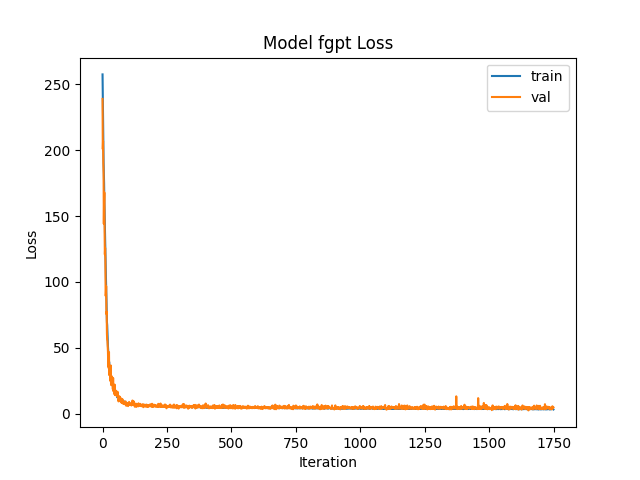}
    \end{subfigure}
    \hfill
    \begin{subfigure}{0.48\columnwidth}
        \centering
        \caption{Normal}
        \includegraphics[width=\textwidth]{./assets/fgpt_exp+hadm_loss.png}
    \end{subfigure}

    \caption{Training Comparison (ELA GPT)}
    \label{fig:mem_comp}
\end{figure}

Notably, the QKV weight matrices of this memory module are inherently pluggable. In principle, the framework can be embedded into any semantic-transformation-based model capable of producing $\Delta X_{k|k-1}$, endowing it with the capacity to learn from internal experience and form qualitative memory. This offers a novel paradigm for LLM training beyond existing approaches such as \textbf{LoRA}\cite{Hu2021lora} and \textbf{Engram}\cite{cheng2026conditional}.

The design of this module draws inspiration from the principles of biological neural memory\cite{polyn2008,zhang2018}, wherein the prefrontal cortex plays a critical role in the contextual integration of memory traces\cite{desousa2026}, as well as from earlier computational memory models such as Neural Turing Machines\cite{Graves2014} and Memory Networks\cite{Weston2014} that established the paradigm of differentiable external memory access.

\section{Experiments}
\label{sec:experiments}
\textbf{To ensure consistent experimental conditions, all attention kernels within a given model variant adopt the same kernel type.}

All models are constructed for controlled ablation studies under a unified experimental setup. The training dataset comprises 129 $\times$ 3,500 samples, totaling 451,500 tokens. We employ the Minimind\cite{Jingyao2026} tokenizer with a vocabulary size of $V = 6{,}400$. The model architecture consists of $L = 4$ Transformer layers, with a model dimension of $d_{\text{model}} = 256$ and $n_{\text{heads}} = 4$ attention heads. A Mixture-of-Experts (MoE) module with $n_{\text{experts}} = 4$ experts is introduced. The total parameter count of the base configuration is 5,838,864. All models are trained for 30 epochs.

We separately train a baseline FA-GPT with standard MoE, as well as ELA-GPT variants equipped with the Hadamard Exp Kernel and the Summation Squared Euclidean Distance Kernel, respectively.

\subsection{No Memory Training Comparison}

\includegraphics[width=\linewidth]{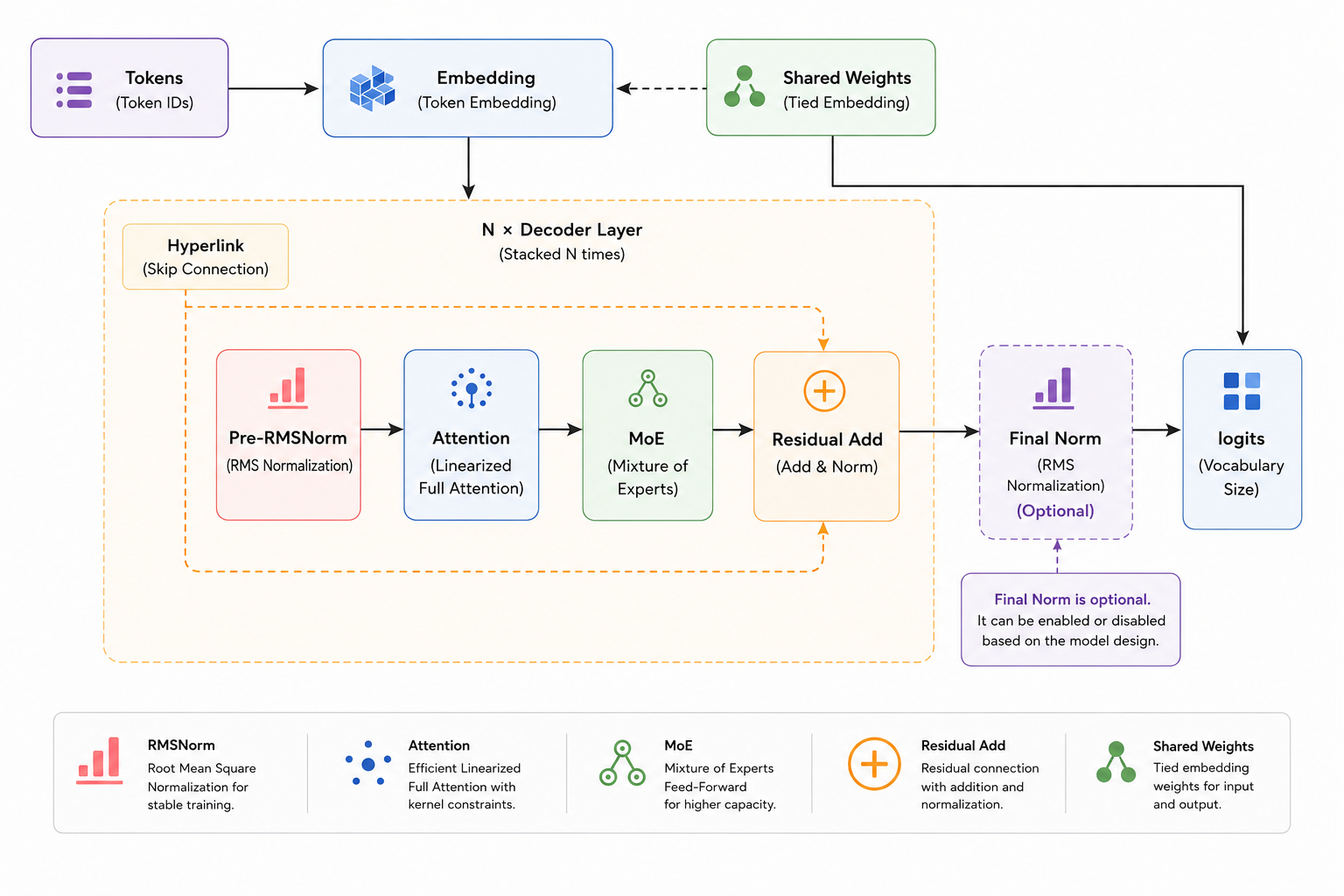}

\begin{figure}[htbp]
    \caption{Training Comparison (Hyper-Link)}
    \centering
    \begin{subfigure}{0.32\columnwidth}
        \centering
        \caption{$\|A_i+B_j\|^2$}
        \includegraphics[width=\textwidth]{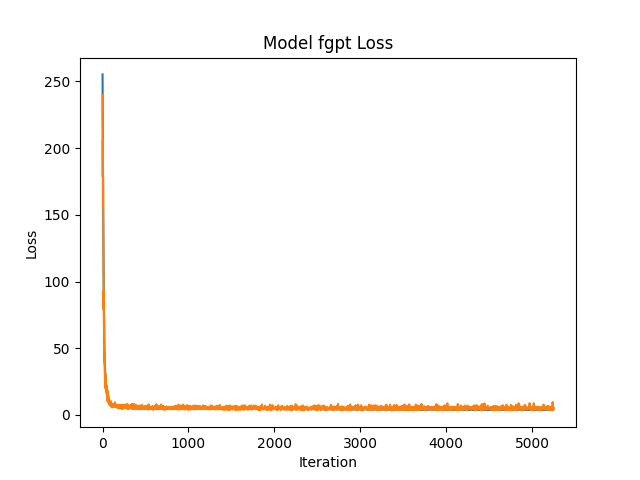}
    \end{subfigure}
    \hfill
    \begin{subfigure}{0.32\columnwidth}
        \centering
        \caption{$exp(A_i)exp(B_j)$}
        \includegraphics[width=\textwidth]{./assets/fgpt_exp+hadm_loss.png}
    \end{subfigure}
    \hfill
    \begin{subfigure}{0.32\columnwidth}
        \centering
        \caption{full}
        \includegraphics[width=\textwidth]{./assets/gpt_loss.png}
    \end{subfigure}
    \label{fig:train_hl}
\end{figure}

\begin{figure}[htbp]
    \caption{Training Comparison (Normal)}
    \centering
    \begin{subfigure}{0.32\columnwidth}
        \centering
        \caption{$\|A_i+B_j\|^2$}
        \includegraphics[width=\textwidth]{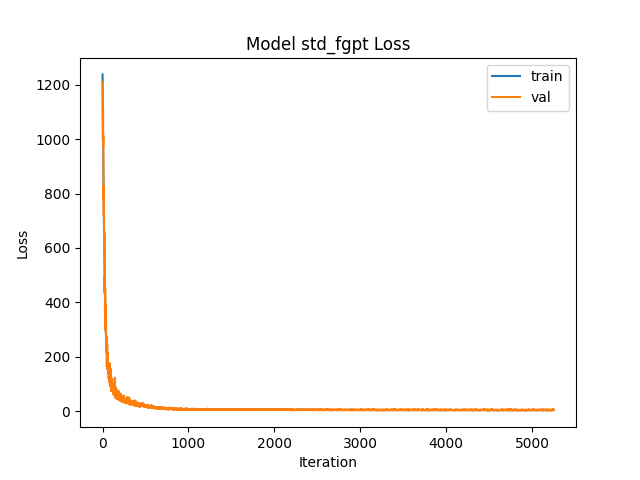}
    \end{subfigure}
    \hfill
    \begin{subfigure}{0.32\columnwidth}
        \centering
        \caption{$exp(A_i)exp(B_j)$}
        \includegraphics[width=\textwidth]{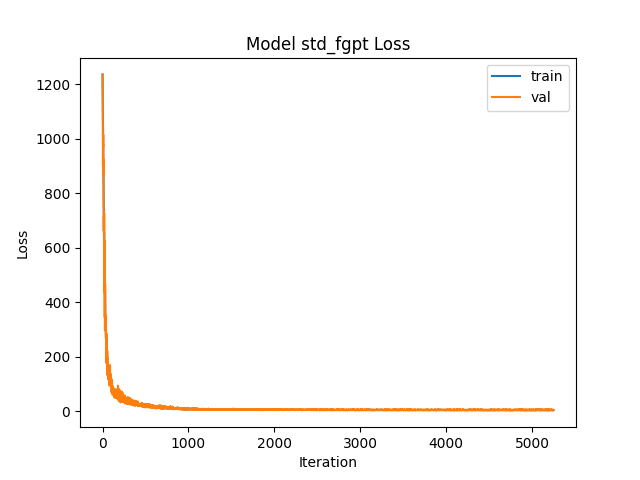}
    \end{subfigure}
    \hfill
    \begin{subfigure}{0.32\columnwidth}
        \centering
        \caption{full}
        \includegraphics[width=\textwidth]{./assets/std_gpt_loss.png}
    \end{subfigure}
    \label{fig:train_normal}
\end{figure}

\autoref{fig:train_hl} and \autoref{fig:train_normal} show that the two model families achieve broadly comparable training performance. Notably, the ELA variants exhibit a modest advantage in resistance to overfitting.

\subsection{Memory Module Ablation}

In this comparative experiment, integrating the Memory module not only accelerates loss convergence but also induces a qualitatively distinct phenomenon: on the vanilla GPT baseline, we observe an abrupt loss drop with a marked inflection point near the 750th training step (counted as global steps; 10 epochs correspond to 1,750 steps).

This is not a spurious ``grokking'' event. Rather, it reflects the point at which the memory module begins to contribute meaningfully, enabling the model to capture latent regularities in the data. Given the relatively modest training corpus in our setup, the increase in non-embedding parameters to 6,624,272 after introducing the memory module allows the model to directly reuse accumulated empirical patterns.

Crucially, when a causal mask is applied to the Memory Query of the vanilla GPT, this abrupt performance improvement vanishes entirely—confirming that the effect is attributable to the bidirectional memory mechanism.

These experimental results collectively demonstrate that the proposed ELA exhibits strong anti-overfitting behavior and generalization capability.
\begin{figure}[htbp]
    \caption{Training Comparison (Hyper-Link \& Memory)}
    \centering
    \begin{subfigure}{0.32\columnwidth}
        \centering
        \caption{$exp(A_i)exp(B_j)$}
        \includegraphics[width=\textwidth]{./assets/fgpt_mem_loss.png}
    \end{subfigure}
    \hfill
    \begin{subfigure}{0.32\columnwidth}
        \centering
        \caption{full}
        \includegraphics[width=\textwidth]{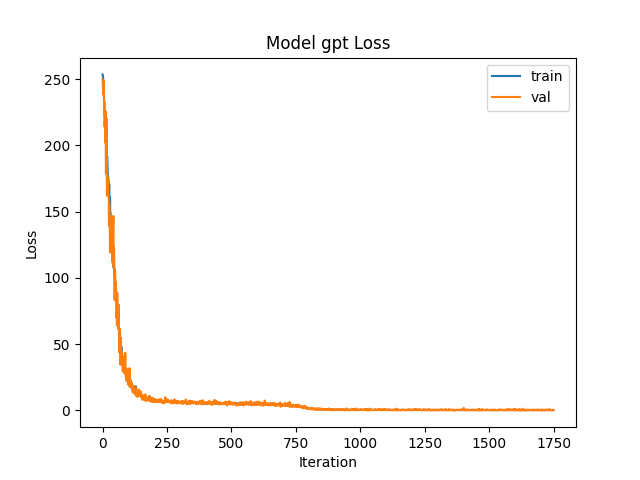}
    \end{subfigure}
    \hfill
    \begin{subfigure}{0.32\columnwidth}
        \centering
        \caption{full(mem-causal)}
        \includegraphics[width=\textwidth]{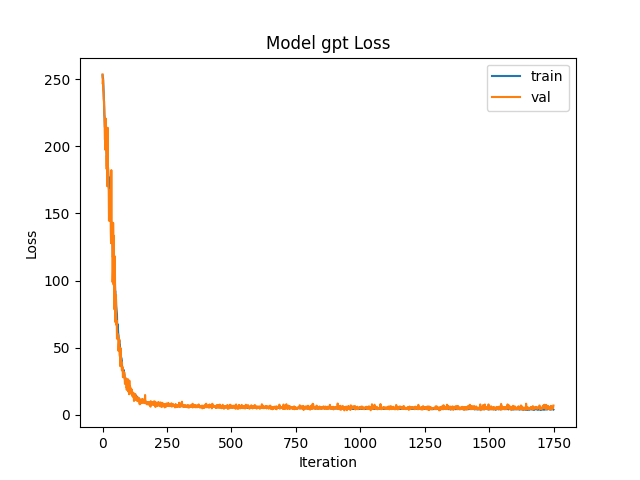}
    \end{subfigure}
    \label{fig:train_hl_mem}
\end{figure}

\subsection{Long-range training}
In long-range training, ELA maintains stable convergence.
\begin{figure}[ht]
  \centering
  \includegraphics[width=\linewidth]{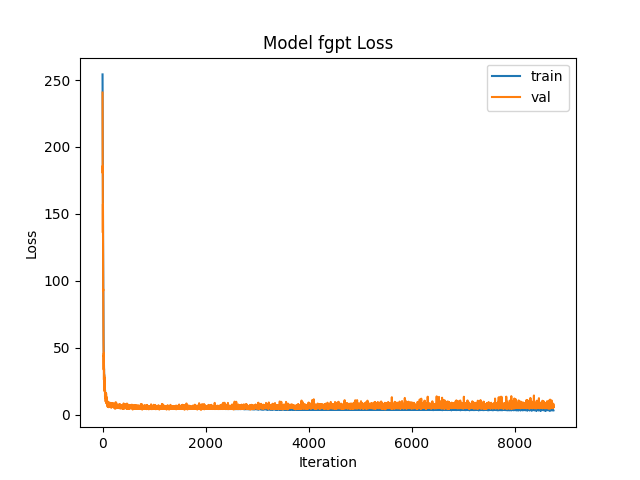}
  \caption{Long-range Training with ELA}
  \label{fig:long}
\end{figure}

\section{Extension to Vision Models}
\label{sec:vision}
We reformulate the deep convolutional layers in YOLO~\cite{YOLOv1,YOLO26,YOLOv8}—the long-standing real-time object detection family—using linear attention, yielding a model with substantially fewer parameters and lower inference latency that delivers strong performance on our benchmarks.

\begin{figure}
    \centering
    \includegraphics[width=\linewidth]{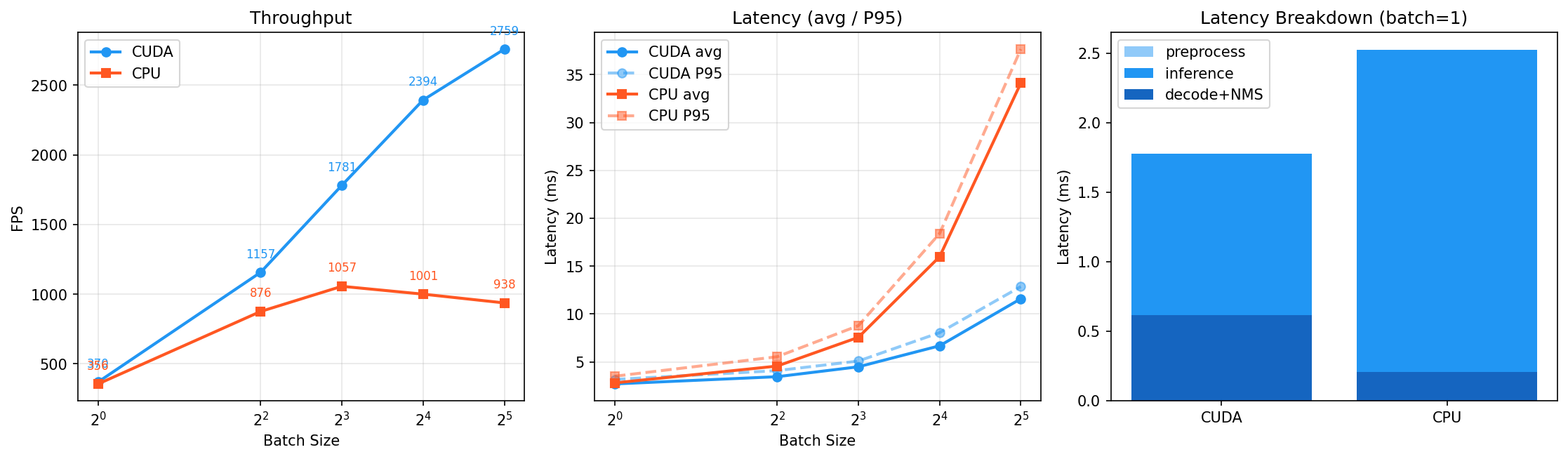}
    \caption{CUDA vs CPU in inference speed}
    \label{fig:CPUvsGPU}
\end{figure}

\begin{figure}
    \centering
    \includegraphics[width=\linewidth]{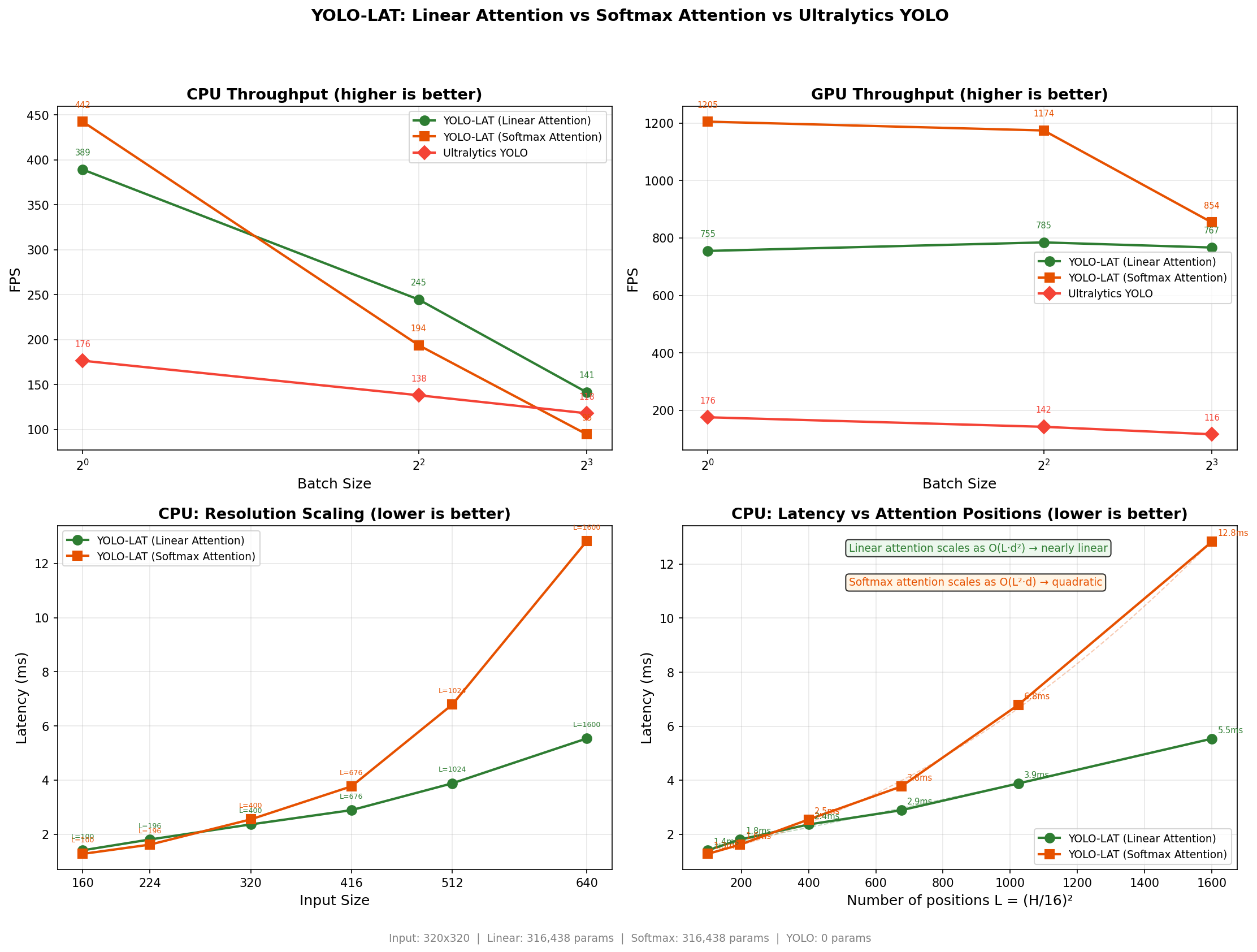}
    \caption{YOLO-LAT vs YOLOv26 in inference speed}
    \label{fig:LATvsYOLOv26}
\end{figure}

\begin{figure}
    \centering
    \includegraphics[width=\linewidth]{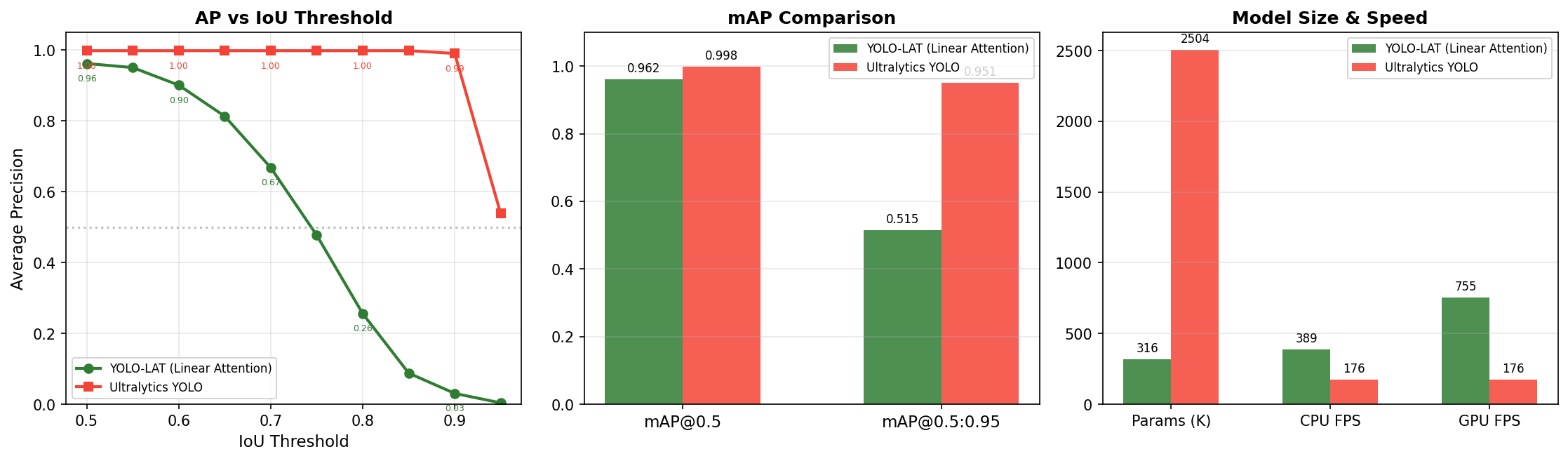}
    \caption{YOLO-LAT vs YOLOv26 in inference accuracy}
    \label{fig:LATvsYOLOv26_acc}
\end{figure}

In head-to-head comparisons, YOLO-LAT achieves \textbf{2.2$\times$ faster inference on CPU} and \textbf{4.3$\times$ faster inference on GPU} relative to vanilla YOLO~\cite{YOLO26}. In terms of accuracy, our method attains competitive performance with \textbf{7.9$\times$ fewer parameters}: YOLO-LAT reaches an mAP@0.5 of 0.962, approaching YOLO’s 0.998. A gap remains, however, in mAP@0.5:0.95 (0.515 versus 0.951), indicating room for improvement in bounding box localization precision.

We attribute this localization gap primarily to the absence of explicit depth cues---an inherent limitation of pure 2D image attention that successive YOLO generations have addressed through architectural specialization: CASC dynamic channel pruning in YOLOv5~\cite{YOLOv5}, the trainable bag-of-freebies paradigm of YOLOv7~\cite{YOLOv7}, and the attention-centric design of YOLO12~\cite{YOLO12}. YOLO-LAT, by contrast, relies solely on intrinsic linear attention without bespoke detection-specific modules---a design choice that deliberately favors \textbf{generality over task-specific optimization}. That YOLO-LAT nonetheless achieves competitive mAP@0.5 while using 7.9$\times$ fewer parameters underscores the representational power of exact linear attention. The localization gap, rather than exposing a fundamental weakness, clarifies the natural boundary between general-purpose attention and domain-specific inductive biases. We regard this as a constructive roadmap: future depth-aware extensions of ELA can reintroduce explicit geometric priors without sacrificing the efficiency and interpretability of exact kernel decomposition. The concurrent work of YOLO-DMA~\cite{YOLO_DMA}, which combines linear attention with deformable convolutions for small-object localization, reinforces this direction. These results collectively demonstrate the viability of extending linear attention~\cite{LinearViT} to visual domains.

To further investigate ELA's impact on Vision Transformer (ViT) architectures, we constructed a simplified object detection model based on FCOS~\cite{tian2019fcos}. Owing to its favorable training dynamics, we conservatively report results after 50 base training epochs to avoid overfitting. The linear-complexity attention of ELA is particularly well-suited to pure vision models: Q, K, and V matrices can be derived natively through convolution operations, after which attention is computed over the full image—a process that mirrors the human visual mechanism of selective target focus. This paradigm aligns with the end-to-end detection philosophy of DETR\cite{Carion2020detr}, while replacing its quadratic self-attention with exact linear attention for improved scalability.

\begin{figure}[ht]
  \centering
  \includegraphics[width=\linewidth]{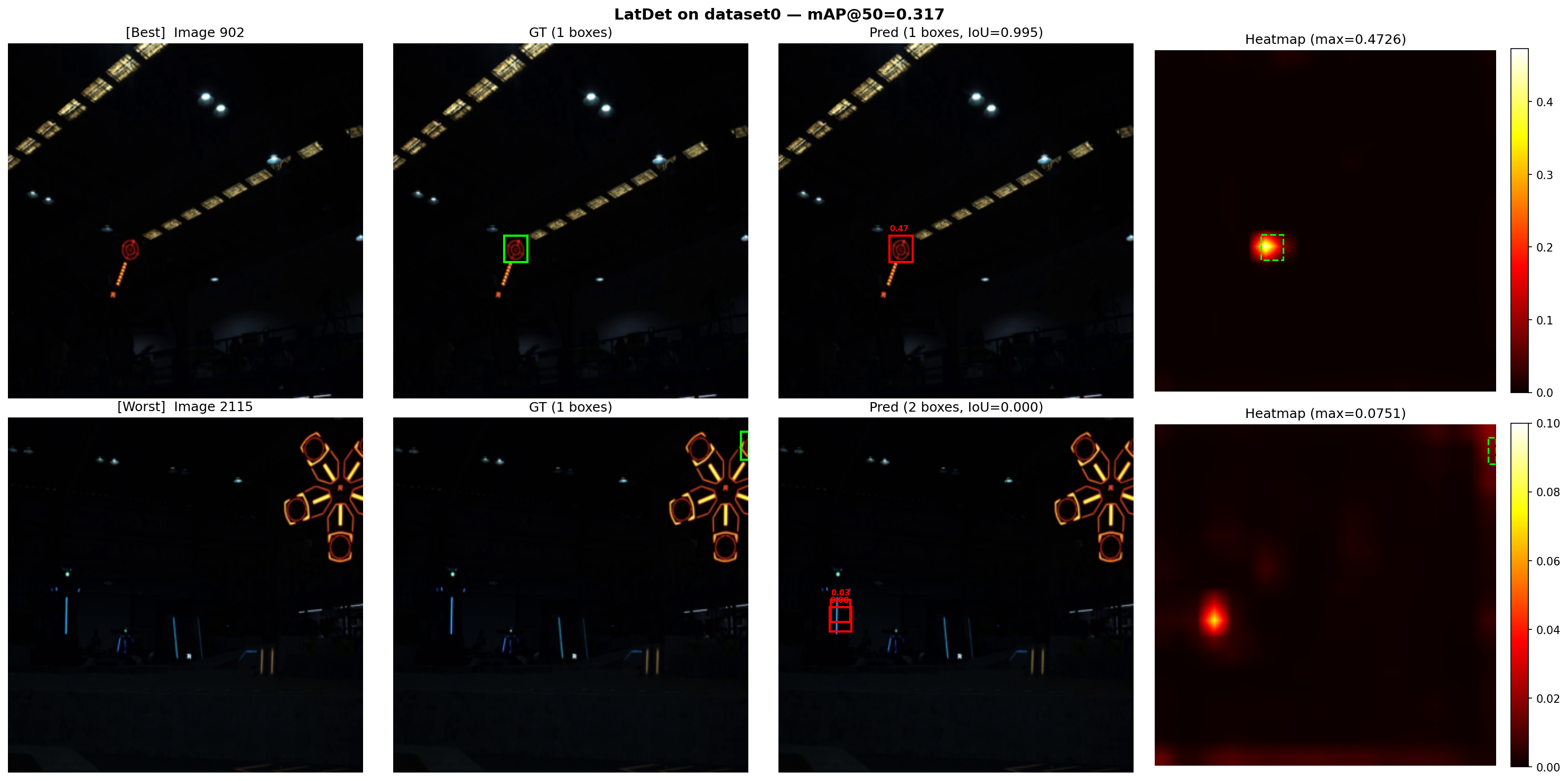}
  \caption{Detection's Attention Vision}
  \label{fig:detection_viz}
\end{figure}

In future work, we plan to incorporate vision-LiDAR fused imagery with native depth cues to train models endowed with intrinsic depth estimation capabilities. More broadly, we view ELA as a mathematical scaffold for \textbf{full-modality world models}: architectures that unify perception, memory, and reasoning within a single, interpretable attention-driven framework. The exact kernel decomposition at the heart of ELA---combined with the pluggable Memory Lobe for qualitative experience accumulation and the label-vector routing for semantic expert coordination---provides the architectural primitives for such unification. Depth-aware vision is a natural first step; the ultimate objective is a model that attends, remembers, and reasons across text, vision, audio, and sensor modalities with mathematically exact, linear-complexity attention.

\section{Future Directions}
Ongoing and planned work extends the framework established in this paper along several promising axes, as summarized in \autoref{tab:future_directions}.
\begin{table}[htbp]
\centering
\caption{Future research directions enabled by Exact Linear Attention.}
\label{tab:future_directions}
\begin{tabular}{@{}p{0.28\columnwidth}p{0.62\columnwidth}@{}}
\toprule
\textbf{Direction} & \textbf{Description} \\
\midrule
Diffusion models & Leverage exact, indefinitely long attention for global perception of fine-grained details, thereby enhancing generation quality in image and video synthesis. \\
\addlinespace
Kernel optimization & Investigate decomposable kernels approximating $e^{xy}$ (the current Hadamard Exp corresponds to $e^{x+y}$), closing the gap to softmax-like behavior while retaining exactness. \\
\addlinespace
Scaling law circumvention & Probe whether the memory module can surpass conventional scaling law constraints\cite{Kaplan2020scaling,Hoffmann2022chinchilla}, achieving stronger performance with fewer parameters via low-rank QK factorization\cite{Hu2021lora}. \\
\addlinespace
State-space models & Bridge exact kernel decomposition with SSMs such as Mamba\cite{Gu2023mamba}, exploring whether the mathematical transparency of ELA's pairwise attention weights can be imparted to structured recurrence for interpretable yet efficient inference. \\
\addlinespace
Full-modality world models & Unify text, vision, audio, and sensor modalities within a single ELA-based architecture, leveraging the Memory Lobe for cross-modal qualitative memory and exact kernel decomposition for interpretable multimodal fusion---a mathematical foundation for general-purpose world understanding. \\
\bottomrule
\end{tabular}
\end{table}
All these endeavors converge on a single thesis: that \textbf{exact, interpretable, linear-complexity attention} is not merely a computational accelerant, but a foundational primitive for the next generation of large-scale AI systems. As models grow toward world-scale understanding---spanning modalities, accumulating qualitative memory, and reasoning over million-token contexts---the mathematical exactness of ELA's kernel decomposition ensures that efficiency gains never come at the cost of representational fidelity. We submit that this principle will prove essential to the construction of general-purpose world models.

\section*{Acknowledgments}
We sincerely thank the anonymous reviewers and the associate editor for their valuable time, rigorous reviews, and insightful constructive feedback. Their professional comments and thoughtful suggestions have greatly helped refine the technical presentation, strengthen the logical framework, and substantially elevate the overall quality of this manuscript.

\end{document}